\documentclass[conference]{IEEEtran}
    \IEEEoverridecommandlockouts

    \usepackage{cite}
    \usepackage{amsmath,amssymb,amsfonts}
    \usepackage{algorithmic}
    \usepackage{graphicx}
    \usepackage{textcomp}
    \usepackage{xcolor}
    \usepackage{booktabs}
    \usepackage{makecell}   
    \usepackage[table]{xcolor}
\definecolor{avgshade}{RGB}{237,235,247}
    \usepackage[labelsep=period]{caption}    
    \usepackage[justification=raggedright, singlelinecheck=false]{caption}
    \def\BibTeX{{\rm B\kern-.05em{\sc i\kern-.025em b}\kern-.08em
        T\kern-.1667em\lower.7ex\hbox{E}\kern-.125emX}}
    \begin{document}
    
    \title{Geometric Encoding for Spatial Reasoning in Vision-Language Models\\
    \thanks{\textsuperscript{*}These authors contributed equally, listed alphabetically.}
    }

    \author{
    \IEEEauthorblockN{Antonio Jun*}
    \IEEEauthorblockA{\textit{Dept. of Computer Science} \\
    \textit{Hunter College}\\
    New York City, United States \\
    antoniojun.nyc@gmail.com}
    \and
    \IEEEauthorblockN{Haoshui Yu*}
    \IEEEauthorblockA{\textit{School of Arts and Sciences} \\
    \textit{New York University}\\
    New York City, United States \\
    hy3635@nyu.edu}
    \and
    \IEEEauthorblockN{Zhengyi Lu}
    \IEEEauthorblockA{\textit{Dept. of Engineering and Computer Science} \\
    \textit{Oakland University}\\
    Rochester, United States \\
    zhengyilu@oakland.edu}
    \and
    \IEEEauthorblockN{Huirong Fu}
    \IEEEauthorblockA{\textit{Dept. of Engineering and Computer Science} \\
    \textit{Oakland University}\\
    Rochester, United States \\
    fu@oakland.edu}
    \and
    \IEEEauthorblockN{Yao Qiang}
    \IEEEauthorblockA{\textit{Dept. of Engineering and Computer Science} \\
    \textit{Oakland University}\\
    Rochester, United States \\
    qiang@oakland.edu}
    }
    
    \maketitle
    
    \begin{abstract}
    Vision-Language Models (VLMs) are far more reliable at recognizing what appears in a video than at reasoning about its spatial and temporal properties, such as metric distances, object dimensions, and consistent object identities across frames. We present Geometric Code, a perception-to-geometry pipeline that computes explicit spatial structure from video and supplies it to VLMs as context to augment reasoning. A perception layer segments and classifies objects and recovers depth, camera pose, and intrinsics from monocular RGB video. A deterministic geometric engine then back-projects, merges, and cleans these outputs into a spatial code, including per-object positions, dimensions, counts, inter-object distances, appearance order, and room geometry. The code is serialized into VLMs' prompts, either alongside the video or replacing it entirely. Specifically, there is no component trained or fine-tuned in our approach. On VSI-Bench, augmenting 2B and 4B open models with the spatial code improves average accuracy by +4.1 points over the frames-only baseline, with the largest gains on numeric estimation tasks such as absolute distance (+24.1 points). The results suggest that explicitly computed geometry, delivered through the language channel, recovers spatial competence that small VLMs cannot extract from pixels alone.
    \end{abstract}
    
    \begin{IEEEkeywords}
    Vision Language Models, Spatial Reasoning, Computer Vision
    \end{IEEEkeywords}
    
    \section{Introduction}
    Humans observe the world as a continuous visual input of 3D environments in our surroundings. When tasked with understanding the physical world around us, we can use a degree of spatial common sense to estimate distance, follow moving objects, and plan an effective route to navigate a complex environment. However, achieving similar performance on spatial reasoning tasks remains challenging for AI models. Prior works aim to address this by training on large-scale spatial data \cite{spatialvlm}, injecting 3D or depth features into the model's architecture \cite{threedllm, spatialrgpt}, or learning encoders that fuse reconstructed geometry with the model \cite{vlm3r}. Recent advancements in Vision Language Models (VLMs), which use visual encoders and adapters to convert image data into language tokens, have improved their spatial reasoning abilities \cite{llava}. While VLMs can effectively reason about what objects are present and their approximate spatial locations, they remain limited in metric distance estimation and cross-frame object coherence due to their lack of explicit geometric understanding \cite{beyondsemantics, viewspatial}.
    
    \begin{figure}[!t]
      \centering
      \includegraphics[width=\columnwidth]{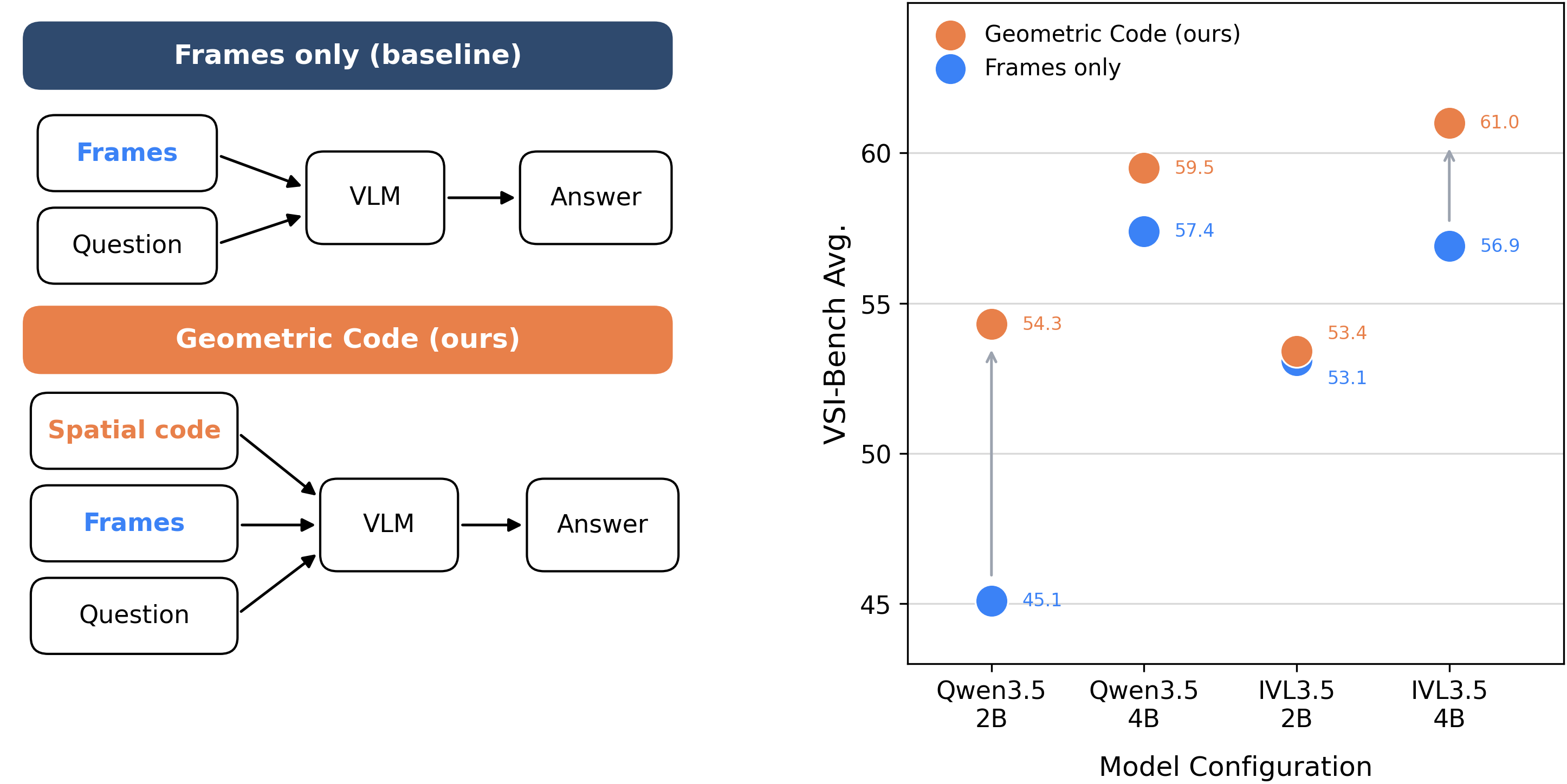}
      \caption{Geometric Code improves spatial reasoning from videos. \emph{Left:} Standard VLMs reason directly from sampled video frames, whereas our framework additionally provides an explicit spatial code derived from geometric analysis of the video frames. \emph{Right:} Average accuracy on VSI-Bench for four open-source VLMs. Augmenting the input with spatial code improves performance over the corresponding frames-only baselines.}
      \label{fig:teaser}
    \end{figure}
    
    
    To address these limitations, we present \textbf{Geometric Code}, a framework that augments VLMs with geometrically derived spatial cues, encoded as spatial code and provided as additional model inputs.
    Our pipeline operates in three stages: 1) A perception layer that segments and classifies each object using SAM-3 \cite{sam3} and recovers geometry through DA-3 \cite{da3}. 2) A geometric engine that receives inputs from the perception layer to compute relevant information for each task. 3) A prompting stage to present spatial code and instructions for VLM reasoning. Unlike prior work, which learns explicit spatial code through a trainable spatial encoder \cite{spatialcode}, our approach combines frozen perception models with a deterministic geometric engine, eliminating the need for training and fine-tuning at any stage. As illustrated in Fig. \ref{fig:teaser}, across four open-source models \cite{Qwen3.5, InternVL} on VSI-Bench \cite{VSI-Bench}, our training-free spatial code augmentation consistently improves average spatial-reasoning accuracy over the frames-only baselines.


    We summarize our contributions as follows:
    \begin{itemize}
        \item We introduce \textbf{Geometric Code}, a training-free framework that augments VLM spatial reasoning at inference time with explicit geometric cues.
        \item We develop a deterministic geometric engine that converts outputs from pretrained perception models into structured spatial codes capturing object-level 3D geometry, requiring no additional training or fine-tuning.
        \item We conduct extensive evaluation and ablation studies on VSI-Bench, showing that Geometric Code provides substantial gains on metric spatial tasks, especially absolute distance.
    \end{itemize}

    \section{Related Works}

    VSI-Bench~\cite{VSI-Bench} established that current VLMs, despite strong
    semantic understanding of video, perform far below humans on spatial
    questions, and that the weakness concentrates in metric estimation.
    Subsequent analyses traced the gap deeper, showing that spatial structure
    is largely absent from VLM representations~\cite{beyondsemantics} and
    that performance collapses when a question requires adopting another
    viewpoint~\cite{viewspatial}. Latest general-purpose
    VLMs~\cite{videollava, llavaonevision, qwen25vl, qwen3} narrow the
    overall gap with scale, but the metric weakness persists, which suggests
    the missing capability is not one that more pixels and parameters
    naturally supply.
    
    One direction is to train the capability in. Early work approached this
    through data, with SpatialVLM~\cite{spatialvlm} synthesizing millions of
    spatial QA pairs for fine-tuning, and through representation, with
    SpatialRGPT~\cite{spatialrgpt} grounding regions in depth while
    3D-LLM~\cite{threedllm} and VLM-3R~\cite{vlm3r} modified architectures to
    consume 3D input directly. More recent methods refine the training
    procedures. For example, SpatialLadder~\cite{spatialladder} builds a curriculum from
    localization up to reasoning. Spatial-MLLM~\cite{spatialmllm} pairs a
    structural encoder with the semantic one. SpaceR~\cite{spacer}
    optimizes spatial reasoning with verifiable rewards. 
    These methods improve benchmark performance, but incur two key costs. They require additional training, and the acquired spatial knowledge remains implicit in model weights, making it difficult to inspect, interpret, or reuse.

    
    Another direction, and the one closest to ours, is to compute the
    geometry outside the model and hand it over as additional spatial information.
    Thinking-with-Spatial-Code~\cite{spatialcode} reconstructs the scene from
    video, serializes it into a spatial code, and shows that a VLM reasons
    better when it can read the geometry it cannot reliably perceive. We
    follow the same channel with a different emphasis. Unlike their pipeline that relies on trained components, our pipeline is training-free. It combines off-the-shelf segmentation and depth models~\cite{sam3, da3} with a deterministic geometric engine. We further isolate the contribution of the spatial code by evaluating frames, code, and their combination across four small open models, revealing what geometry contributes with and without the visual input it summarizes.

    \begin{figure*}[t]
      \centering
      \includegraphics[width=1\textwidth]{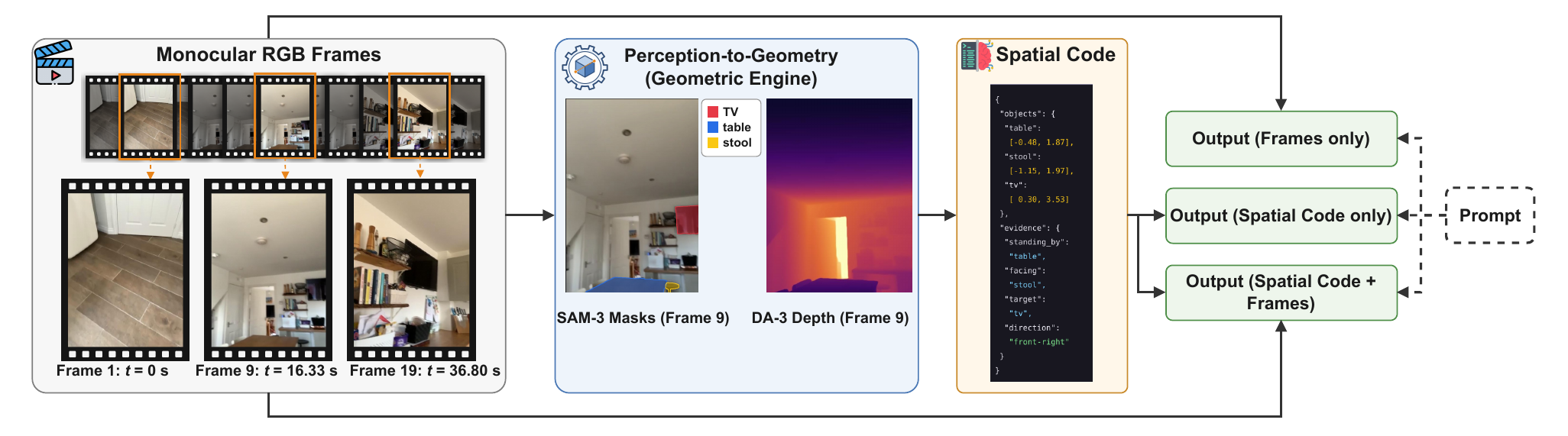}
      \caption{\textbf{Overview of the Geometric Code.} Monocular RGB frames are processed
      by the perception stage, where SAM-3 produces per-object segmentation
      masks and DA-3 estimates per-frame metric depth. A geometric engine
      back-projects these outputs into a spatial code, a JSON summary of
      per-object 3D positions and precomputed spatial evidence. Performance
      is evaluated under three VLM input conditions: frames only, spatial
      code only, and both.}
      \label{fig:pipeline}
    \end{figure*}
        
    \section{Methodology}

    \subsection{Overview}
    
    Fig.~\ref{fig:pipeline} illustrates the overall workflow of our Geometric Code framework. Given a video and a question, we first sample monocular video frames as the visual input. These frames are then processed by a perception layer, where Segment Anything Model (SAM-3)~\cite{sam3} identifies and segments objects and Depth Anything (DA-3)~\cite{da3} estimates metric depth. The resulting object masks and depth maps are passed to a deterministic geometric engine, which reconstructs object-level 3D information and computes task-relevant spatial relationships. The geometric information is then converted into a structured spatial code that summarizes object positions and spatial evidence in a compact textual form. This code is provided to VLMs together with the video frames for spatial reasoning. 

    \subsection{Perception Layer}
    \label{sec:perception}
    
    The intention of the perception layer is to identify and provide relevant raw metrics as inputs into the geometric engine. The perception layer is composed of two vision models, i.e., SAM-3 for object segmentation and DA-3 for depth approximation. 
    
    \subsubsection{Segmentation} 

    For object perception, we apply SAM-3 directly to monocular video frames sampled at 6~fps using an open-vocabulary object list. For each VSI-Bench dataset, the vocabulary is constructed as the union of all object categories appearing in the released annotations and is fixed for all scenes in that dataset, ensuring that no scene-specific object information is introduced. Given a frame and the shared vocabulary, SAM-3 identifies all object instances whose matching scores exceed its default confidence threshold and returns an instance-level binary mask for each detection, including multiple instances belonging to the same semantic category. Specifically, each frame is processed independently, and the instance indices produced by SAM-3 are local to that frame. We therefore do not need to assume that mask identities are temporally consistent or that the same index corresponds to the same object across different frames.

    It is important to note here that we use segmentation masks rather than bounding boxes because masks more closely follow object contours and therefore reduce the inclusion of irrelevant background pixels. This is particularly important for the subsequent geometric stage, where depth and spatial measurements are computed from pixels associated with each object. Background contamination can otherwise bias the estimated object geometry. 

    
    

    \subsubsection{Depth Approximation} Although SAM-3 can effectively create 2D masks for objects, it still lacks depth dimensions. Thus, we supplement the SAM-3 model with DA-3 to provide per-frame forward depth, camera intrinsics, camera pose, and per-pixel confidence estimates. DA-3 divides each frame into patches of pixels and predicts depth from learned monocular and multi-view cues. Forward depth is a pixel's distance measured along the camera's viewing direction, rather than the straight-line distance to the camera itself. Since patches are attended to jointly across all frames in a processing window, DA-3 can estimate the position and orientation relevant to camera pose from how the matched image regions shift between frames. Across all frames available to a scene, we use DA-3 streaming with a chunk size of 120 and overlap of 60 frames to provide consistent geometric estimates across frames. Since SAM-3 and DA-3 operate over the same set of frames, each mask can be paired with the corresponding depth, pose, and camera intrinsics. This alignment provides the geometric basis for projecting each segmented object from image space into a consistent 3D representation used by the following geometric engine.

    \subsection{Geometric Engine}
    \label{sec:engine}
    
    \subsubsection{Overview} 
    The geometric engine deterministically converts the outputs of the perception layer into spatial code, ensuring that identical inputs always produce identical outputs without any learned components. Using a shared set of intermediate geometric representations, it then computes the task-specific quantities required for spatial reasoning.

    \subsubsection{Back-projection}
    The first step of the geometric engine fuses the segmentation and depth streams into a unified 3D representation. For each frame $f$ and each detected object instance, we consider all pixels $(u,v)$ inside the corresponding SAM-3 mask. Each pixel is paired with its forward depth $z = D_f(u,v)$, obtained from DA-3. This associates every segmented object pixel with both an image-plane location and a depth measurement.  Using the camera intrinsics, each masked pixel is then back-projected from image space into the 3D camera coordinate system~\cite{backprojection}. Specifically, the pixel location $(u,v)$ and its depth $z$ are converted into a 3D point $\mathbf{x}_c$ as:
    \begin{equation}
    \mathbf{x}_c =
    \left(
    \frac{(u-c_x)z}{f_x},
    \frac{(v-c_y)z}{f_y},
    z
    \right).
    \label{eq:unproject}
    \end{equation}
    Here, $c_x$ and $c_y$ denote the principal point, while $f_x$ and $f_y$ are the focal lengths along the horizontal and vertical image axes. The first two coordinates recover the lateral position of the pixel relative to the camera, while $z$ represents its forward distance along the camera viewing direction.
    
    Because the camera coordinate system changes as the camera moves, the resulting 3D points cannot yet be directly compared across frames. We therefore transform each point into a shared world coordinate system using the estimated camera pose for frame $f$. Given camera rotation $\mathbf{R}_f$ and translation $\mathbf{t}_f$, the transformation is formulated as:
    \begin{equation}
    \mathbf{x}_w = \mathbf{R}_f\mathbf{x}_c + \mathbf{t}_f,
    \label{eq:world}
    \end{equation}
    where $\mathbf{x}_w$ denotes the corresponding 3D point in world coordinates. Expressing points from all frames in the same coordinate system allows geometric measurements to be compared consistently over time. Thus, for each object mask, the transformed pixels form a 3D point cloud that captures the object’s visible surface in the current frame. These object-level point clouds provide a unified geometric representation from which the engine derives downstream spatial attributes, including positions, distances, sizes, and spatial relationships.


    The point cloud is then cleaned by removing outliers from samples so as to retain only the values that most reliably represent the geometric shapes. So, any points that are on the outskirts or potentially involve depth estimations from the background that would obscure the accurate estimation of this object are discarded. Numerically, any point $p$ with average distance $\bar{d}(p)$ to its $k$ closest neighbors is compared to cloud-wide mean $\mu$ and standard deviation $\sigma$ of the cloud. That point is discarded if it constitutes an outlier by a threshold of $\bar{d}(p) > \mu + 2\sigma$. 
    
    
    
    
    
    

        \subsubsection{Canonical Frame} Currently, the world frame which unifies all objects is taken from the first frame where the most objects appear. That frame is still defined by the camera which may not necessarily align with the precise coordinates and axis of the room. This transformation intends to create that alignment by mapping out the orientation of the floor and the upward vectors. First, the engine recovers the vertical by fitting a plane to a confidence-filtered sample of scene points with RANSAC \cite{ransac}, exploiting the two properties that characterize the floor: it is large and flat, and almost nothing lies below it. Each hypothesis is a candidate floor plane, parameterized by a pair $(\mathbf{n}, d)$. The unit normal $\mathbf{n}$ fixes the plane's orientation: it is the direction perpendicular to the plane, intended to be the upward vector. The offset $d$ fixes the plane's position along that direction. Together they define the plane as the set of points satisfying
    \begin{equation}
        \label{eq:signeddist}
        s(\mathbf{p}) = \mathbf{n} \cdot \mathbf{p} + d = 0.
    \end{equation}
     The value $s(\mathbf{p})$ is the signed perpendicular distance of
    $\mathbf{p}$ from the hypothesis plane: magnitude encodes distance and sign encodes which side of the plane it lies on. Since $\|\mathbf{n}\| = 1$, $s$ carries metric units, so one physical threshold applies fairly to every hypothesis.
     
    Both properties of the floor can now be checked directly on $s$. Since measured points are noisy, lying on the plane is relaxed to a slab of tolerance $\varepsilon = 5$ cm: of the $N$ sampled points, $N_0$ lie inside the slab and $N_+$, $N_-$ stand clear of it on either side. Over 300 hypotheses, the floor is the plane maximizing
    \begin{equation}
    \label{eq:floorscore}
    J(\mathbf{n},d) = N_0 \cdot \frac{\max(N_+,\, N_-)}{N},
    \end{equation}
    the number of points on the plane times the fraction of \emph{all} sampled points that stand clear of it on its dominant side. The two factors enforce the two floor properties in turn: many measured points must lie on the surface, rejecting empty parallel planes that intersect no points, and the points that do not must be concentrated on a single side, rejecting planes that slice through the interior of the scene and split the sample between them. We take the maximum of $N_+$ and $N_-$ because the labeling of the two sides is arbitrary: $(\mathbf{n},d)$ and $(-\mathbf{n},-d)$ describe the same plane with $N_+$ and $N_-$ exchanged, and the maximum makes $J$ invariant to that choice. Once the winning plane is chosen we fix its orientation so that the cameras lie on the positive side: since they film from above the floor, if they read $s < 0$ we replace $(\mathbf{n}, d)$ with $(-\mathbf{n}, -d)$, so that $\mathbf{n}$ points away from the floor and $\mathbf{b}_3 = \mathbf{n}$ becomes the vertical.
    
    If no plane survives, the engine falls back to the scene's axis of least extent. The horizontal axes $\mathbf{b}_1, \mathbf{b}_2$ are the world axes projected onto the floor, correcting only tilt, and the floor level is the second percentile of object-point heights along $\mathbf{b}_3$.

    \subsubsection{Instance Assembly}\label{sec:assembly} 
    
    After aligning the world coordinate frame with the room axes, the next step is to determine whether detections observed in different frames correspond to the same object. Because each frame is segmented independently, repeated observations of the same object are initially treated as separate detections. Before computing object-level geometric quantities, the engine must therefore associate detections across frames and distinguish repeated observations from genuinely distinct object instances. Each detection is represented by a 3D bounding box whose extent is robustly estimated using the 2nd and 98th percentiles of its constituent points.
    

    Two detections belonging to the same semantic class are eligible for merging if their 3D bounding volumes overlap and they are never observed simultaneously in the same frame. The co-visibility constraint follows from the fact that two same-class detections that appear in the same frame must correspond to distinct objects and therefore cannot be assigned to the same instance. Importantly, merging is performed at the group level rather than independently for individual detection pairs. For every candidate merge, the combined sets of frame indices associated with the two groups must remain disjoint. This prevents a detection from being indirectly merged, through a chain of intermediate associations, with another detection that was co-visible with it. The procedure therefore avoids introducing additional tuned similarity thresholds.
    

    This group-level constraint is necessary because naive pairwise merging can violate the co-visibility requirement through transitivity. For example, suppose detections $A$ and $C$ satisfy the merge criteria, and detections $A$ and $B$ also satisfy them individually. A purely transitive procedure could place $A$, $B$, and $C$ into the same object instance even if $B$ and $C$ were observed together in one frame. To prevent this inconsistency, merge conditions are propagated across the entire candidate groups. Thus, after $A$ and $C$ are merged, a subsequent attempt to merge their group with $B$ is rejected if any detection in the existing group is co-visible with $B$.
    

    The output of this step is a set of consolidated object instances, each associated with its estimated 3D position, spatial extent, and occupied region in the room-aligned world coordinate system. These instances provide a consistent object-level representation for subsequent computation of geometric and spatial relationships.
    
    \label{subsubsec:Object_position}
    
    \subsubsection{Task -- Absolute Distance} 
    The absolute distance between two object instances, $A$ and $B$, is defined as the minimum Euclidean distance between any pair of points in their respective 3D point clouds, formally:
    \begin{equation}
        d(A, B) = \min_{\mathbf{a} \in A,\, \mathbf{b} \in B}
        \lVert \mathbf{a} - \mathbf{b} \rVert.
        \label{eq:dist}
    \end{equation}
    This definition measures the closest surface-to-surface separation between the two objects, rather than the distance between their centroids, and therefore better reflects their true spatial relationships.
        
    \subsubsection{Task -- Relative Distance} 
    Following instance assembly, each detected object instance is represented in the room-aligned world coordinate frame by a 3D bounding box whose extent is defined by the 2nd and 98th percentiles of its constituent points along each canonical axis. This representation provides an approximate object position and spatial extent for every consolidated instance.

    Using these object representations, the engine computes a pairwise distance matrix over detected object classes, where each entry is obtained from Eq.~\ref{eq:dist} and reported to centimeter precision. In addition to the absolute pairwise distances, each row records the relative ordering of all other classes by proximity. This ranked representation reduces the reasoning burden on downstream models, which no longer need to compare multiple numerical entries in the distance matrix to determine relative closeness. Instead, the nearest class to a given query object can be directly identified as rank 1, with subsequent classes ordered by increasing distance.
    
    

    \subsubsection{Task -- Object Size} 
    Object size is estimated from the instance's best observation, defined as the single frame whose mask and corresponding point cloud receive the highest accuracy score. Using a single well-observed view avoids combining partial or occluded observations across multiple frames, which could distort the estimated dimensions. For the selected frame, the engine fixes the vertical axis to the room's gravity direction and determines the two horizontal axes using principal component analysis (PCA) on the floor-projected object points. This allows the horizontal axes to align with the object's dominant orientation rather than with the room axes. The extent along each axis is then computed as the span between the 2nd and 98th percentiles of the projected points, consistent with the robust trimming strategy used during instance assembly. The resulting three spans represent the object's estimated side lengths. These dimensions are ordered from largest to smallest, and the longest dimension is retained as the final size estimate. Because VSI-Bench evaluates object size using the object's longest dimension, the engine reports only the maximum span.

    \subsubsection{Task -- Room Size} 
    
    Floor area is estimated at the scene level rather than from individual object point clouds. High-confidence depth points aggregated across all frames are first projected onto the floor plane of the canonical coordinate frame (Eq.~\ref{eq:floorscore}). The projected points are then trimmed using central percentiles to suppress outliers and stray reconstruction points. Floor area is computed as the area enclosed by an alpha shape fitted to the remaining points, with $\alpha = 2$, selected to match the density of our reconstructed point clouds. 
    In addition to the scalar floor-area estimate, the engine constructs an explicit room outline from the same floor-projected points. The floor plane is discretized into $10$\,cm grid cells, small gaps in the occupancy map are closed, and the resulting boundary is traced and simplified to obtain a compact representation of the room footprint.
    
    
        
    \subsubsection{Task -- Object Count} 
    Once instance assembly is complete, the spatial engine reports the number of detected object instances for each semantic class. The resulting count is inherently constrained by detection recall: any object that is not successfully segmented in any frame cannot be recovered during instance assembly and therefore cannot contribute to the final count. Consequently, counting errors are more likely to manifest as undercounting than overcounting.

    \subsubsection{Task -- Relative Direction} 
    Building on the approximate object positions and spatial extents obtained from the preceding stages, the geometric engine retains a consolidated representation of each unique object's location and occupied region. In parallel, the camera-path estimation provides interval-level camera motion, which is used to infer the camera orientation associated with selected sub-samples of frames. These object-centric geometric features and camera-motion cues are then integrated and distilled into a compact spatial code for the VLM to process.
    
    
    \subsubsection{Task -- Appearance Order} 
    Classes are ordered according to their first appearance in the input sequence, using the earliest frame index retained for each consolidated instance. During instance assembly, subsequent detections of the same physical object are associated with its initial observation, such that the resulting instance preserves the frame ID of its first appearance. The engine therefore outputs only the ordered sequence of object classes, rather than the underlying frame indices or timestamps, because the benchmark evaluates relative appearance order rather than the exact time of observation.
    
    \begin{table*}[t]
        \centering
        \caption{Video spatial reasoning results on VSI-Bench \cite{VSI-Bench}.
        Comparison of four open-source MLLMs under the frames-only baseline,
        spatial code only, and combined settings. The upper blocks report results
        from prior work as compiled by Thinking-with-Spatial-Code \cite{spatialcode}. Comparability is approximate due to evaluation-setup differences. Rows under
        \textit{Geometric Code} are our own evaluations, with ``Frames Only'', ``Code Only'',  and ``Frames + Code'' denoting our input configurations. \textbf{Bold} indicates the best result per column across all methods and \underline{underline} the second best.}
        \label{tab:main}
        \small
        \setlength{\tabcolsep}{5pt}
        \renewcommand{\arraystretch}{1.12}
        \begin{tabular}{ll>{\columncolor{avgshade}}ccccccccc}
        \toprule
         & & \multicolumn{9}{c}{\textbf{VSI-Bench}} \\
        \cmidrule(lr){3-11}
        \textbf{Methods} & \textbf{Size} & \textbf{Avg.} & \makecell{\textbf{Obj.}\\\textbf{Count}} & \makecell{\textbf{Abs.}\\\textbf{Dist.}} & \makecell{\textbf{Obj.}\\\textbf{Size}} & \makecell{\textbf{Room}\\\textbf{Size}} & \makecell{\textbf{Rel.}\\\textbf{Dist.}} & \makecell{\textbf{Rel.}\\\textbf{Dir.}} & \makecell{\textbf{Route}\\\textbf{Plan}} & \makecell{\textbf{Appear.}\\\textbf{Order}} \\
        \midrule
        Human Level & -- & 79.2 & 94.3 & 47.0 & 60.4 & 45.9 & 94.7 & 95.8 & 95.8 & 100 \\
        \midrule
        \multicolumn{11}{l}{\textit{Spatial-centric MLLMs}} \\
        \hspace{1em}SpatialLadder~\cite{spatialladder} & 3B & 44.8 & 62.1 & 35.3 & 61.9 & 41.4 & 45.6 & 46.4 & 27.3 & 38.5 \\
        \hspace{1em}Spatial-MLLM~\cite{spatialmllm} & 4B & 46.3 & 66.6 & 38.0 & 63.6 & 35.4 & 40.4 & 48.2 & 32.9 & 44.3 \\
        \hspace{1em}SpaceR~\cite{spacer} & 7B & 41.5 & 44.5 & 24.7 & 53.5 & 37.3 & 41.9 & 46.1 & 29.3 & 54.8 \\
        \hspace{1em}Thinking w/ Spatial Code~\cite{spatialcode} & 4B & 57.0 & 58.3 & 39.0 & \textbf{73.0} & 52.4 & 57.8 & 55.9 & \textbf{38.7} & 63.9 \\
        \hspace{1em}Thinking w/ Spatial Code + 2D box~\cite{spatialcode} & 4B & \underline{60.0} & \textbf{95.2} & 60.7 & 50.8 & 33.1 & \underline{62.0} & \textbf{87.1} & 32.5 & 59.0 \\
        \midrule
        \multicolumn{11}{l}{\textit{Open-source MLLMs}} \\
        \hspace{1em}LLaVA-Video~\cite{videollava} & 7B & 35.6 & 48.5 & 14.0 & 47.8 & 24.2 & 43.5 & 42.4 & 34.0 & 30.6 \\
        \hspace{1em}LLaVA-Video~\cite{videollava} & 72B & 40.9 & 48.9 & 22.8 & 57.4 & 35.3 & 42.4 & 36.7 & 35.0 & 48.6 \\
        \hspace{1em}LLaVA-OneVision~\cite{llavaonevision} & 7B & 32.4 & 47.7 & 20.2 & 47.4 & 12.3 & 42.5 & 35.2 & 29.4 & 24.4 \\
        \hspace{1em}LLaVA-OneVision~\cite{llavaonevision} & 72B & 40.2 & 43.5 & 23.9 & 57.6 & 37.5 & 42.5 & 39.9 & 32.5 & 44.6 \\
        \hspace{1em}Qwen2.5-VL~\cite{qwen25vl} & 7B & 32.3 & 32.8 & 18.1 & 43.8 & 31.7 & 38.0 & 37.4 & 28.3 & 27.9 \\
        \hspace{1em}Qwen3-VL~\cite{qwen3} & 8B & 55.0 & 52.1 & 44.7 & 60.4 & 43.1 & 56.6 & 56.3 & \underline{38.1} & \textbf{70.2} \\
        \hspace{1em}Qwen3-VL~\cite{qwen3} & 4B & 52.8 & 53.1 & 46.3 & 63.4 & 48.0 & 53.3 & 49.9 & 37.1 & 58.9 \\
        \hspace{1em}Qwen3-VL + 2D box~\cite{qwen3} & 4B & 54.5 & 66.6 & 42.3 & 57.8 & 40.8 & 56.9 & 52.7 & 37.6 & 66.9 \\
        \midrule
        \multicolumn{11}{l}{\textit{Geometric Code} (ours)} \\
        \hspace{1em}Qwen3.5-2B (Frames Only) & 2B & 45.1 & 58.1 & 34.2 & 62.3 & 49.4 & 46.1 & 40.7 & 34.5 & 28.2 \\
        \hspace{1em}Qwen3.5-2B (Code Only) & 2B & 45.9 & 41.4 & 65.8 & 58.9 & \textbf{65.1} & 34.5 & 39.7 & 26.8 & 23.1 \\
        \hspace{1em}Qwen3.5-2B (Frames + Code) & 2B & 54.3 & 41.5 & 61.3 & 56.7 & 60.0 & 59.7 & 49.3 & 29.9 & 59.4 \\
        \midrule
        \hspace{1em}Qwen3.5-4B (Frames Only) & 4B & 57.4 & 61.6 & 37.6 & \underline{68.6} & 55.5 & 59.6 & 58.9 & 31.4 & \underline{67.5} \\
        \hspace{1em}Qwen3.5-4B (Code Only) & 4B & 56.5 & 41.4 & 65.8 & 60.0 & \textbf{65.1} & 60.3 & 63.4 & 32.5 & 40.6 \\
        \hspace{1em}Qwen3.5-4B (Frames + Code) & 4B & 59.5 & 40.4 & \underline{67.1} & 59.9 & \textbf{65.1} & \textbf{62.5} & 63.5 & 32.0 & 62.3 \\
        \midrule
        \hspace{1em}InternVL3.5-2B (Frames Only) & 2B & 53.1 & 68.8 & 39.4 & 65.0 & 50.5 & 54.9 & 45.5 & 35.6 & 55.5 \\
        \hspace{1em}InternVL3.5-2B (Code Only) & 2B & 53.8 & 41.4 & 65.4 & 59.2 & \textbf{65.1} & 61.0 & 44.9 & 29.9 & 49.0 \\
        \hspace{1em}InternVL3.5-2B (Frames + Code) & 2B & 53.4 & 44.1 & 66.3 & 59.5 & 57.3 & 43.8 & 52.2 & \textbf{38.7} & 50.8 \\
        \midrule
        \hspace{1em}InternVL3.5-4B (Frames Only) & 4B & 56.9 & \underline{70.3} & 43.2 & 68.0 & \underline{65.0} & 53.5 & 47.5 & 36.6 & 67.3 \\
        \hspace{1em}InternVL3.5-4B (Code Only) & 4B & 56.8 & 41.4 & \textbf{67.3} & 50.9 & \textbf{65.1} & 58.6 & \underline{64.7} & \underline{38.1} & 52.9 \\
        \hspace{1em}\textbf{InternVL3.5-4B (Frames + Code)} & 4B & \textbf{61.0} & 53.4 & \textbf{67.3} & 60.4 & \textbf{65.1} & 60.0 & 64.4 & \textbf{38.7} & 61.5 \\
        \bottomrule
        \end{tabular}
    \end{table*}

    \subsubsection{Task -- Route Planning} 
    \label{Camera_pathing}
    The camera trajectory is sampled at 5-second intervals and represented as a sequence of waypoints, each encoding the camera's horizontal position and heading. The heading is obtained by projecting the camera's forward direction onto the floor plane. These waypoints provide explicit motion cues for questions involving the observer's movement through the scene, which are not captured by the object-centric spatial code alone. The engine does not explicitly compute bearings or routes; instead, these tasks are resolved by the VLM through reasoning over the emitted camera trajectory, object positions, and room outline.

    \subsection{Prompting}\label{sec:prompting}
    The spatial code is serialized in JSON format and preceded by a concise natural-language legend, with measurement units explicitly embedded in the values, e.g., ``1.18 meters''. In the code-only setting, this representation replaces the video input entirely. For numeric questions, the model is instructed to return a single word or short phrase and is decoded greedily with a maximum generation length of 16 tokens and reasoning disabled. For multiple-choice questions, the answer options are included in the prompt and the model is allocated a 1024-token reasoning budget. A fixed suffix, ``The correct option is:'', is then appended to elicit a final committed response. Specifically, greedy decoding is used in all settings, ensuring deterministic outputs.
    

    \section{Results}

    \subsection{Experimental Setup}
    
    Our framework generates spatial hints in two main stages. First, the perception layer applies SAM-3 for open-vocabulary object detection and segmentation and DA-3 for depth and camera geometry estimation. Second, a deterministic geometric engine aligns and fuses these outputs in a common coordinate system, computes task-relevant spatial quantities, and converts them into structured spatial hints that are provided to the VLM together with the original prompt.
    The experiments focus on 2B- and 4B-parameter variants of Qwen3.5 and InternVL3.5, enabling us to evaluate whether the observed trends generalize across different VLM families and model scales. Each model is evaluated under three input configurations: frames only, using 32 uniformly sampled frames as the baseline; code only, using only the generated spatial code; and frames with code, where the spatial code is provided together with the video frames. For metric estimation tasks, the geometric engine deterministically computes the required quantities, and the VLM mainly retrieves and reports the corresponding values from the spatial code. Consequently, performance on these tasks can be nearly identical across models and primarily reflects the quality of the geometric engine rather than the reasoning capability of the VLM. In contrast, tasks such as object count and object size require the VLM to reason over the spatial code, the video frames, or both, making their performance more dependent on the model itself, as shown in Table \ref{tab:main}.

    \begin{table}[t]
        \centering
        \caption{Input-configuration ablation on VSI-Bench, reported as average accuracy. Values in parentheses ($\Delta$) indicate changes relative to the Frames baseline for the corresponding model. \textbf{bold} denotes the best-performing configuration for each model. \textit{Overall Avg.}\ is the unweighted mean across the four models.}
        \label{tab:ablation_avg}
        \small
        \setlength{\tabcolsep}{4.5pt}
        \renewcommand{\arraystretch}{1.18}
        \begin{tabular}{lccc}
        \toprule
        \textbf{Model} & \textbf{Frames Only} & \textbf{Code Only} & \textbf{Frames+Code} \\
        \midrule
        Qwen3.5-2B     & 45.1 & 45.9 \,{\scriptsize($+0.8$)} & \textbf{54.3} \,{\scriptsize($+9.2$)} \\
        Qwen3.5-4B     & 57.4 & 56.5 \,{\scriptsize($-0.9$)} & \textbf{59.5} \,{\scriptsize($+2.1$)} \\
        InternVL3.5-2B & 53.1 & \textbf{53.8} \,{\scriptsize($+0.7$)} & 53.4 \,{\scriptsize($+0.3$)} \\
        InternVL3.5-4B & 56.9 & 56.8 \,{\scriptsize($-0.1$)} & \textbf{61.0} \,{\scriptsize($+4.1$)} \\
        \midrule
        \textit{Overall Avg.} & 53.1 & 53.3 \,{\scriptsize($+0.1$)} & \textbf{57.1} \,{\scriptsize($+3.9$)} \\
        \bottomrule
        \end{tabular}
    \end{table}
    \subsection{Primary Findings}

    Across the evaluated models, Geometric Code provides the largest benefit when combined with visual frames. As shown in Table~\ref{tab:main}, the code-only setting slightly outperforms the frames-only baseline for both 2B models, improving Qwen3.5-2B from 45.1 to 45.9 and InternVL3.5-2B from 53.1 to 53.8, while offering little or no gain for the 4B models. In contrast, combining frames with spatial code improves average accuracy for all four models, with gains ranging from +0.3 to +9.2 points. The largest improvement occurs for Qwen3.5-2B, which increases from 45.1 to 54.3, while InternVL3.5-4B improves from 56.9 to 61.0 and achieves the best overall performance.
    
    The gains are particularly pronounced on geometry-intensive tasks. \textit{Absolute Distance} improves substantially across all models, reaching 67.3 for InternVL3.5-4B, while \textit{Relative Direction} also benefits consistently from the added geometric information. However, improvements are not uniform across tasks. \textit{Object Count} and \textit{Object Size} generally decline when spatial code is introduced, and \textit{Appearance Order} exhibits model-dependent behavior. These results suggest that explicit geometric cues are most effective for tasks requiring metric and directional reasoning, while appearance- and recognition-oriented tasks continue to rely heavily on information contained in the original visual frames.
    

    \subsection{Ablation Studies} 
\begin{figure}[t]
        \centering
        \includegraphics[width=\columnwidth]{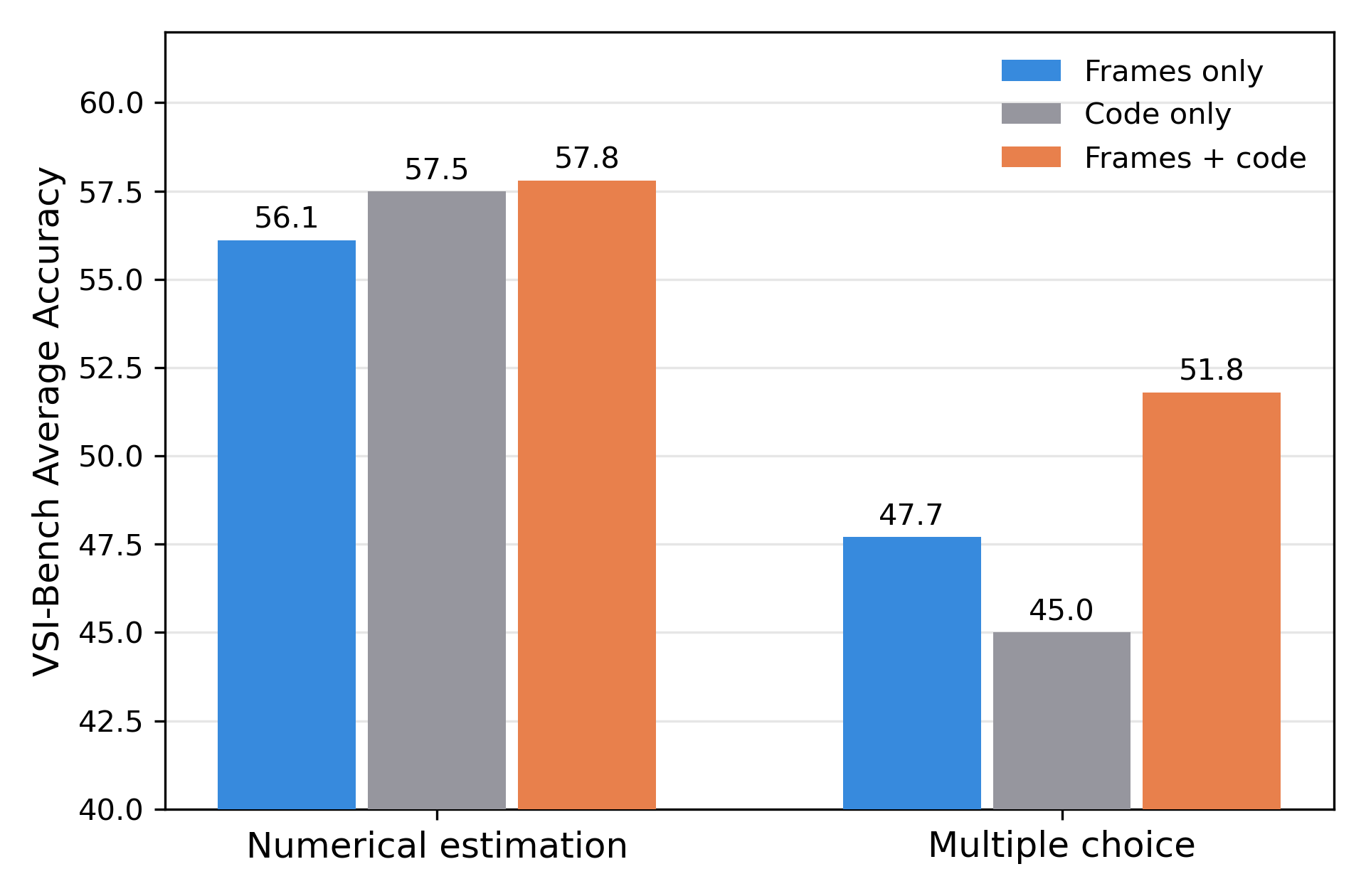}
        \caption{Effect of input configuration on VSI-Bench, averaged across four model configurations and grouped by numerical estimation and multiple choice tasks. Spatial code alone improves numerical estimation but reduces multiple choice accuracy, while combining frames with spatial code achieves the best performance on both task types.}
        \label{fig:na_vs_mc}
    \end{figure}

    We conduct three ablations over the input given to the VLM: the effect of each input configuration on average accuracy (Table~\ref{tab:ablation_avg}), a decomposition by answer format (Fig.~\ref{fig:na_vs_mc}), and a task-level analysis over the full breakdown (Table~\ref{tab:main}).

    \subsubsection{Input Configuration} 
    Code only has little effect on overall performance. Averaged across the four models, it increases accuracy only marginally from 53.1 to 53.3, with model-level changes ranging from $-0.9$ to $+0.8$ points relative to the frames-only baseline. Combining frames with spatial code produces a much clearer benefit. The Frames+Code configuration improves the overall average from 53.1 to 57.1, corresponding to a mean gain of $+3.9$ points over frames only. It is the best configuration for three of the four models, yielding gains of $+9.2$ for Qwen3.5-2B, $+2.1$ for Qwen3.5-4B, and $+4.1$ for InternVL3.5-4B. The largest improvement occurs for Qwen3.5-2B, where Frames+Code also exceeds Code only by 8.4 points. 
    Overall, spatial code is most effective as a complementary geometric representation rather than a replacement for visual frames. The strongest gains arise when explicit geometry and raw visual evidence are provided together, suggesting that the two modalities contribute distinct and complementary information to spatial reasoning.
    
    \begin{figure*}[t]
      \centering
      \includegraphics[width=\textwidth]{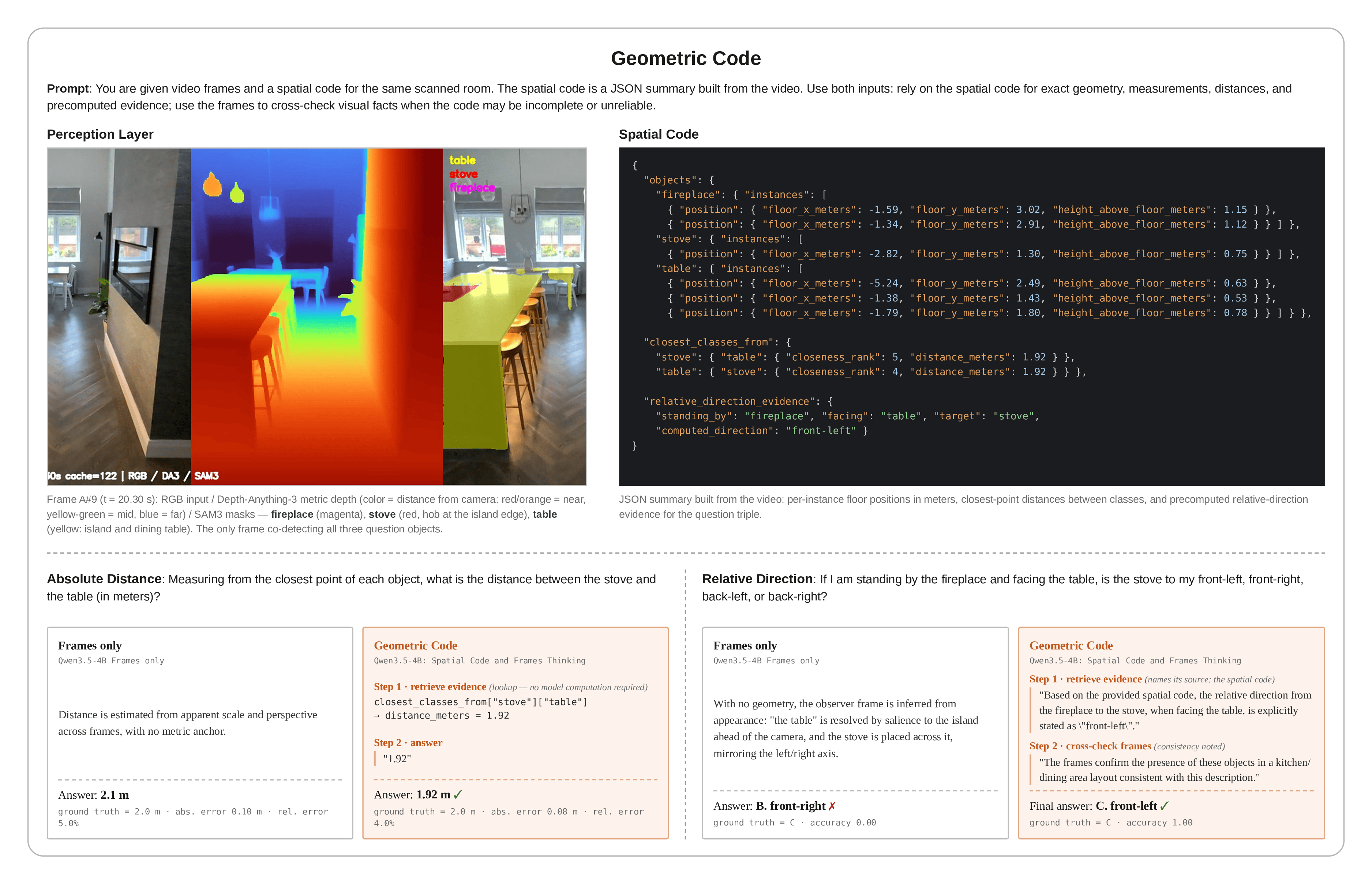}
      \caption{A case study sample built from an ARKitScenes scene with Qwen3.5-4B thinking. It showcases one question from each task: numeric-estimation Absolute Distance and multiple-choice Relative Direction. We provide reasoning for frames only in contrast to Geometric Code to showcase how Qwen3.5-4B reasons using both input configurations.}
      \label{fig:case_study}
    \end{figure*}

    \subsubsection{Task-Type Decomposition} 
    Fig.~\ref{fig:na_vs_mc} decomposes the overall results by answer format and reveals a clear difference between numerical-estimation and multiple-choice tasks. For numerical estimation, Code only improves average accuracy from 56.1 to 57.5, while Frames+Code further increases it to 57.8. In contrast, Code only reduces multiple-choice accuracy from 47.7 to 45.0, whereas combining code with frames raises performance to 51.8. Thus, spatial code is particularly effective for quantitative estimation but is less sufficient for tasks that require broader visual context. Providing both representations yields the strongest performance in both categories, suggesting that explicit geometric measurements and raw visual evidence provide complementary information for VLM reasoning.


    \subsubsection{Task-Level Analysis}
    \label{sec:task_level} 
    Frames and spatial code exhibit complementary strengths across task categories. As shown in Table~\ref{tab:main}, frames only generally performs better on \textit{Object Count}, \textit{Object Size}, and \textit{Appearance Order}, whereas code only provides substantial gains on geometry-driven tasks such as \textit{Absolute Distance} and \textit{Room Size}. This distinction does not align cleanly with the benchmark's numerical-versus-multiple-choice split. \textit{Object Count} and \textit{Object Size} are both numerical tasks, yet their performance declines with code only, suggesting that the key factor is whether the answer can be directly derived from the reconstructed geometry rather than the output format itself. The clearest example is \textit{Absolute Distance}, where code-only raises accuracy from 34.2--43.2 under frames only to 65.4--67.3 across the four models. 

    In contrast, \textit{Object Count} shows the largest degradation, dropping from 58.1--70.3 with frames to 41.4 with code only. Other reasoning-dependent tasks, including \textit{Relative Distance}, \textit{Relative Direction}, and \textit{Route Planning}, show more model-dependent behavior, indicating that they require a mixture of geometric information and visual-semantic reasoning for optimal performance. Consistent with this idea, raising model size from 2B to 4B improves reasoning-dependent scores while leaving the precomputed estimation tasks, \textit{Object Count} and \textit{Room Size}, largely unchanged; varying the InternVL3.5 and Qwen3.5 model families similarly leaves estimation tasks untouched while changing reasoning-dependent tasks, although more unpredictably in contrast to model size.

    
    \section{Conclusion}

    We introduced Geometric Code, a training-free framework that converts monocular video into structured spatial code to augment VLM spatial reasoning. The framework combines off-the-shelf segmentation and depth models with a deterministic geometric engine that recovers object-level 3D information and expresses it as explicit textual cues. We evaluate Geometric Code across multiple VLM families, model scales, and input configurations on VSI-Bench. Providing spatial code together with video frames improves average accuracy by 3.9 points across four open models, while our strongest configuration reaches 61.0 average accuracy, surpassing previously reported results at a comparable model scale. Task-level analysis further shows that spatial code is particularly effective for geometry-driven reasoning, especially metric estimation, whereas visual frames remain important for tasks that depend on appearance and broader scene context. These results demonstrate that explicit geometric representations can complement visual inputs and improve spatial reasoning without requiring additional training or fine-tuning.


    Two directions naturally follow from this work. First, our results suggest a complementary division of research between deterministic geometry and learned spatial reasoning. The perception-to-geometry pipeline is strongest on metric estimation, whereas fine-tuned approaches~\cite{spatialcode, spatialrgpt} are better suited to reasoning-intensive multiple-choice tasks. A promising extension is therefore to combine these strengths, for example by using parameter-efficient adaptation such as LoRA~\cite{lora} to train VLMs to reason more effectively over Geometric Code. Such a hybrid approach could preserve the precision and interpretability of deterministic measurements while improving performance on tasks that require more complex semantic reasoning. Second, Geometric Code can serve as a general scene representation beyond benchmark question answering. It already encodes object geometry, room structure, and camera trajectory, which are natural inputs for learned world models. Coupling this explicit geometric representation with predictive world models~\cite{vjepa2} could enable reasoning about future states, physical interactions, and temporal dynamics. A key challenge, however, is that most current world models encode spatial knowledge implicitly in latent or generative representations, making it difficult to extract and align their internal geometry with an explicit spatial interface such as Geometric Code.


    \end{document}